\documentclass[letterpaper, 10 pt, conference]{ieeeconf}  

\IEEEoverridecommandlockouts                              

\usepackage[numbers]{natbib}
\usepackage{graphicx}
\usepackage{amsmath}
\usepackage{hyperref}
\usepackage{algpseudocode}
\usepackage{algorithm}
\usepackage[algo2e]{algorithm2e}
\usepackage{multirow}
\usepackage[normalem]{ulem}
\usepackage{booktabs}
\usepackage{gensymb}
\usepackage{lipsum}
\usepackage{amsmath,amsfonts,bm}
\usepackage{caption}
\usepackage{subcaption}

\def\eqref#1{equation~\ref{#1}}
\def\1{\bm{1}}

\DeclareMathAlphabet{\mathsfit}{\encodingdefault}{\sfdefault}{m}{sl}
\SetMathAlphabet{\mathsfit}{bold}{\encodingdefault}{\sfdefault}{bx}{n}

\newcommand{\xxnote}[3]{}
\ifx\hidenotes\undefined
  \renewcommand{\xxnote}[3]{\color{#2}{#1: #3}}
\fi

\hypersetup{
    colorlinks=true,
    linkcolor=blue,
    filecolor=magenta,      
    citecolor=blue,
    urlcolor=blue,
}

\usepackage[font={small}]{caption}
\renewcommand\thefigure{\arabic{figure}}
\newcommand{\method}{\textsc{Holo-Dex}}
\newcommand{\website}{\url{https://holo-dex.github.io/}}

\title{\LARGE \bf
Holo-Dex: Teaching Dexterity with Immersive Mixed Reality
}

\author{
Sridhar Pandian Arunachalam\\
New York University
\and 
Irmak Güzey \\
New York University
\and 
Soumith Chintala \\
Meta AI
\and 
Lerrel Pinto\\
New York University
\thanks{Correspondence to \texttt{sridhar@nyu.edu}.}
}

\newboolean{include-notes}
\setboolean{include-notes}{true}
\newcommand{\BE}[1]{\ifthenelse{\boolean{include-notes}}%
{\textcolor{green}{\textbf{BE: #1}}}{}}

\begin{document}

\makeatletter
\let\@oldmaketitle\@maketitle%
\renewcommand{\@maketitle}{\@oldmaketitle%
    \centering
    \includegraphics[width=\linewidth]{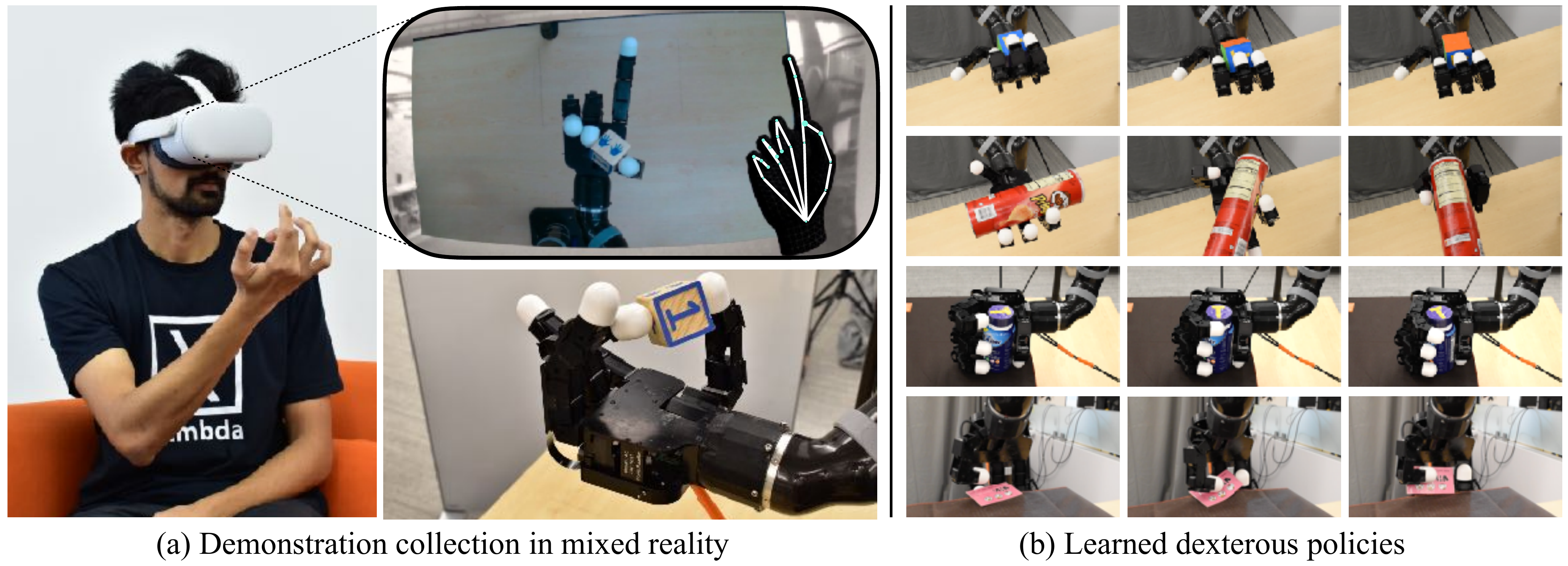}
    \captionof{figure}{We present \method{}, a framework that (a) collects high-quality demonstration data by placing human teachers in an immersive mixed reality world, and then (b) learns visual policies from a handful of these demonstrations to solve dexterous manipulation tasks.}
    \label{fig:intro}
}
\makeatother

\maketitle
\thispagestyle{empty}
\pagestyle{empty}

%%%%%%%%%%%%%%%%%%%%%%%%%%%%%%%%%%%%%%%%%%%%%%%%%%%%%%%%%%%%%%%%%%%%%%%%%%%%%%%%
\begin{abstract}
A fundamental challenge in teaching robots is to provide an effective interface for human teachers to demonstrate useful skills to a robot. This challenge is exacerbated in dexterous manipulation, where teaching high-dimensional, contact-rich behaviors often require esoteric teleoperation tools. In this work, we present \method{}, a framework for dexterous manipulation that places a teacher in an immersive mixed reality through commodity VR headsets. The high-fidelity hand pose estimator onboard the headset is used to teleoperate the robot and collect demonstrations for a variety of general-purpose dexterous tasks. Given these demonstrations, we use powerful feature learning combined with non-parametric imitation to train dexterous skills. Our experiments on six common dexterous tasks, including in-hand rotation, spinning, and bottle opening, indicate that \method{} can both collect high-quality demonstration data and train skills in a matter of hours. Finally, we find that our trained skills can exhibit generalization on objects not seen in training. Videos of \method{} are available on \website.

\end{abstract}

\setcounter{figure}{1}

%%%%%%%%%%%%%%%%%%%%%%%%%%%%%%%%%%%%%%%%%%%%%%%%%%%%%%%%%%%%%%%%%%%%%%%%%%%%%%%%
%%%%%%%%%%%%%%%%%%%%%%%%%%%%%%%%%%%%%%%%%%%%%%%%%%%%%%%%%%%%%%%%%%%%%%%%%%%%%%%%
\section{Introduction}

Learning-based methods have had a transformational effect in robotics on a wide range of domains from manipulation~\cite{chi2022iterative, tekden2020belief}, locomotion~\cite{gangapurwala2022rloc, ma2022combining, smith2022walk}, and aerial robotics~\cite{zhang2016learning, gandhi2017learning, hwangbo2017control}. Such methods often produce policies that input raw sensory observations and output robot actions. This circumvents challenges in developing state-estimation modules, modeling object properties and tuning controller gains, which requires significant domain expertise. Even with the steep progress in robot learning, we are still long way off from dexterous robots that can solve arbitrary robot tasks akin to methods in game play~\cite{berner2019dota, silver2016mastering}, text generation~\cite{guo2018long, huang2021sarg} or few-shot vision~\cite{dosovitskiy2020image, gidaris2018dynamic}.

To understand what might be missing in robot learning, we need to ask a central question: How do we collect training data for our robots? One option is to collect data on the robot through self-supervised data collection strategies. While this results in robust behaviors~\cite{pinto2016supersizing, levine2016learning, singh2019end, vecerik2019practical}, they often require extensive real-world interactions in the order of thousands of hours even for relatively simple manipulation tasks~\cite{dulac2020empirical}. An alternate option is to train on simulated data and then transfer to the real robot (\textit{Sim2Real}). This allows for learning complex robotic behaviors multiple orders of magnitude faster than on-robot learning~\cite{Openai2018,Openai2019}. However, setting up simulated robot environments and specifying simulator parameters often requires extensive domain expertise~\cite{tobin2017domain, jakobi1995sim}.

A third, more practical option to collect data is by asking human teachers to provide demonstrations~\cite{pomerleau1989alvinn,abbeel2004apprenticeship}.  Robots can then be trained to quickly imitate the demonstrated data. Such imitation methods have recently shown promise in a variety of challenging dexterous manipulation problems~\cite{rajeswaran2017learning,DexPilot,arunachalam2022dexterous}. However, there lies a fundamental limitation in most of these works -- collecting high-quality demonstration data for dexterous robots is hard! They either require expensive gloves~\cite{glovereview}, extensive calibration~\cite{DexPilot}, or suffer from monocular occlusions~\cite{arunachalam2022dexterous}.

In this work, we present \method{}, a new framework to collect demonstration data and train dexterous robots. It uses VR headsets (e.g. Quest 2) to put human teachers in an immersive virtual world. In this virtual world, the teacher can view a robotic scene from the eyes of a robot, and control it using their hands through inbuilt pose detectors. \method{} allows humans to seamlessly provide robots with high-quality demonstration data through a low-latency observational feedback system. \method{} offers three benefits: (a) Compared to self-supervised data collection methods, it allows for rapid training without reward specification as it is built on powerful imitation learning techniques; (b) Compared to Sim2Real approaches, our learned policies are directly executable on real robots since they are trained on real data; (c) Compared to other imitation approaches, it significantly reduces the need for domain expertise since even untrained humans can operate VR devices.

We experimentally evaluate \method{} on six dexterous manipulation tasks that require performing complex, contact-rich behavior. These tasks range from in-hand object manipulation to single-handed bottle opening. 
Across our tasks, we find that a teacher can provide demonstrations at an average of $60s$ per demonstration  using \method{}, which is $1.8\times$ faster than prior work in single-image teleoperation~\cite{arunachalam2022dexterous}. On $4/6$ tasks, \method{} can learn policies that achieve $>90\%$ success rates. Surprisingly, we find that the dexterous policies learned through \method{} can generalize on new, previously unseen objects. 

In summary, this work presents \method{}, a new framework for dexterous imitation learning with the following contributions. First, we demonstrate that high-quality teleoperation can be achieved by immersing human teachers in mixed reality through inexpensive VR headsets. Second, we experimentally show that the demonstrations collected by \method{} can be used to train effective, and general-purpose dexterous manipulation behaviors. Third, we analyze and ablate \method{} over various decisions such as the choice of hand tracker and imitation learning methods. Finally, we will release the mixed reality API, demonstrations collected, and training code associated with \method{} on \website.

%%%%%%%%%%%%%%%%%%%%%%%%%%%%%%%%%%%%%%%%%%%%%%%%%%%%%%%%%%%%%%%%%%%%%%%%%%%%%%%%
\section{Related Work}
Our framework builds upon several important works in robot learning, imitation learning, teleoperation and dexterous manipulation. In this section, we briefly describe prior research that is most relevant to ours.

\subsection{Methodologies for Teaching Robots}
There are several approaches one can take to teach robots. Reinforcement Learning (RL) \cite{Kaelbling1996,lillicrap2015continuous,yarats2021mastering} can train policies to maximize rewards while collecting data in an automated manner. This process often requires a roboticist to specify the reward function along with ensuring safety during self-supervised data collection~\cite{levine2016learning,pinto2016supersizing}. Furthermore, such approaches are often sample-inefficient and might require extensive simulation training for optimizing complex skills. 

Simulation to Real (\textit{Sim2Real}) approaches focus on training RL policies in simulation, followed by transferring to the real robot~\cite{tobin2017domain,sadeghi2016cad2rl,pinto2017asymmetric}. Such a methodology of robot training has received significant success owing to the improvements in modern robot simulators. Sim2Real still requires significant human involvement as every task needs to be carefully modeled in the simulator. Moreover, even during training special techniques are required to ensure that the resulting policies can transfer to the real robot~\cite{Openai2019, wu2019learning, rao2020rl, qin2022one}.  

Imitation learning approaches focus on training policies from demonstrations provided by an expert. Behavior Cloning (BC) is an offline technique that trains a policy to imitate the expert behavior in a supervised manner~\cite{pomerleau1989alvinn,florence2021implicit,bojarski2016end,young2020visual}. Recently, non-parametric imitation approaches have shown promise in learning from fewer demonstrations~\cite{rob_lwr,pari2021surprising,arunachalam2022dexterous}. Another set of imitation learning is Inverse Reinforcement Learning (IRL)~\cite{IRL_gt, abbeel2004apprenticeship, haldar2022watch}. Here, a reward function is inferred from demonstrations, followed by using  RL to optimize the inferred reward. While \method{} is geared towards offline imitation, the demonstrations we collect are compatible with IRL approaches as well.

\begin{figure*}[ht!]
    \centering
    \includegraphics[width=\linewidth]{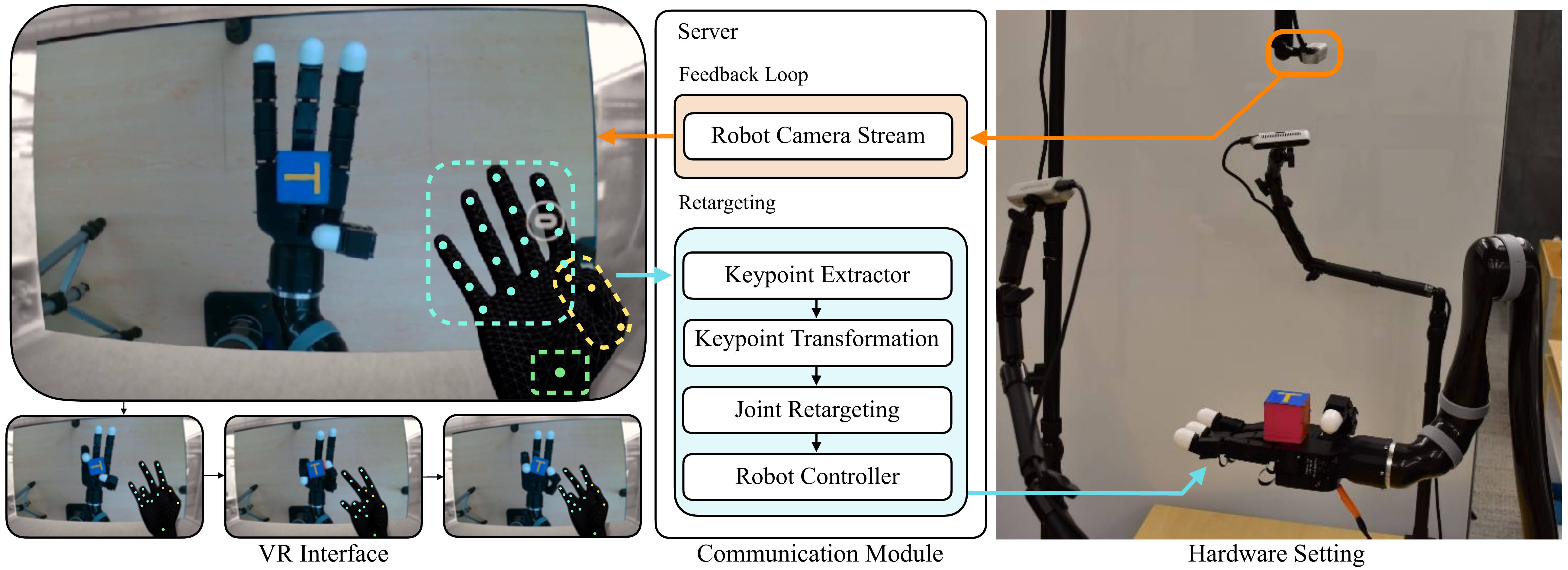} 
    \caption{Overview of \method{}'s teleoperation module. Given a hand pose in the VR interface, the controller streams the keypoint data to the robot's server which transforms and retargets the human hand key points to the Allegro Hand. Visual feedback of the teleoperated hand is then provided back to the VR Headset for real-time feedback.}
    \label{fig:aug2real}
    \vspace{-0.1in}
\end{figure*}

\subsection{Dexterous Teleoperation Frameworks}

To effectively use imitation learning for dexterous manipulation we need to obtain accurate hand poses from a human teacher. There are several approaches to gather demonstrations for dexterous tasks. 
Using a custom glove to measure a user's hand movements such as CyberGlove \cite{glovereview, Kumar2015} or Shadow Dexterous Glove \cite{li2020mobile} has been a popular solution. However, although such gloves have high accuracy, they can be expensive and require significant calibration effort.
Vision-based hand pose detectors have shown promise for dexterous tasks. Some examples include using multiple RGBD~\cite{DexPilot}, single depth ~\cite{li2019vision}, RGB ~\cite{arunachalam2022dexterous}, and RGBD~\cite{qin2022one} images. However, such methods either require custom calibration procedures~\cite{DexPilot} or suffer from occlusion-related issues when using single cameras~\cite{arunachalam2022dexterous}.
Recently, a new generation of VR headsets has enabled advanced multi-camera hand pose detection \cite{megatrack} that gave promising results in \cite{hentschel2022steady, salvato2022predicting}.
This enhancement provides a  robust solution that is significantly cheaper compared to CyberGlove and requires little calibration. While VR tools have been used to collect demonstrations~\cite{gharaybeh2019telerobotic, zhang2018deep} for low-dimensional end-effector control, \method{} shows that the VR headsets can be used for high-dimensional control in augmented reality. Concurrent to our work, Radosavovic et al.~\cite{MVP} also show that hand tracking from VR can be used to teleoperate robot hands albeit without using mixed reality. 

\subsection{Dexterous Manipulation}

Due to its high-dimensional action space, learning complex skills with dexterous multi-fingered robot hand has been a longstanding challenge~\cite{old_dex_man_1, old_dex_man_2, old_dex_man_3, old_dex_man_4}. Model-based RL and control approaches have demonstrated significant success on tasks such as spinning objects and in-hand manipulation~\cite{Kumar2016,Nagabandi2019}. Similarly, model-free RL approaches have shown that Sim2Real can enable impressive skills such as in-hand cube rotation and Rubik's cube face turning~\cite{Openai2018, Openai2019}. However, both learning approaches requires hand-designing reward functions along with system identification~\cite{Kumar2016} or task-specific training procedures~\cite{Openai2019}. Coupled with long training times, often requiring weeks~\cite{Openai2018,Openai2019}, they make dexterous manipulation difficult to scale for general tasks.

To address the poor sample efficiency of prior learning-based methods, several works have looked at imitation learning~\cite{Rajeswaran2018,radosavovic2021state}. Here, given a handful of demonstrations, simulated policies can be trained in a few hours. More recently, such imitation-based approaches have shown success on real robot hands~\cite{arunachalam2022dexterous}. \method{} takes this idea further by improving the teaching process and demonstrating its utility on a variety of in-hand manipulation tasks.

%%%%%%%%%%%%%%%%%%%%%%%%%%%%%%%%%%%%%%%%%%%%%%%%%%%%%%%%%%%%%%%%%%%%%%%%%%%%%%%%
\section{Background on Visual Imitation Learning}
To understand the imitation learning framework used in \method{}, we first formalize and describe important background work in self-supervised learning and non-parametric imitation. Together, these enable efficient imitation learning from high-dimensional visual observations.

\subsection{Visual Self-Supervised Learning} \label{sec:ssl_background}
Self-Supervised Learning (SSL) focuses on obtaining low-dimensional embeddings $z$ from high-dimensional observations $o$~\cite{ericsson2022self}. Operationally, the observations (e.g. RGB images) are fed into an encoder $f_{\theta}$, where $\theta$ denotes the weights of a parametric deep network. While there are several methods to train $f_{\theta}$, the central principle in many works is to predict one `view' of the observation given a different `view' of the same observation. One example of such a learning scheme is data-augmented SSL. Here, the observation $o$ is augmented by applying visual augmentations such as color jitter or random grayscale. Given two augmented views of this observation $o^1$ and $o^2$, the corresponding embeddings would be $z^1 \equiv f_{\theta}(o^1)$ and $z^2 \equiv f_{\theta}(o^2)$. The training objective for $f_{\theta}$ amounts to maximizing the mutual information between the two embeddings $I(z^1, z^2)$.

To optimize this objective, we use the BYOL~\cite{grill2020bootstrap} training scheme, which amounts to predicting $z^2 \leftarrow g_{\phi}(z^1)$ through a small deep model $g_{\phi}$ called the `projector'. This scheme for learning embeddings has had significant success in a variety of domains ranging from computer vision, audio processing, and robotics. Given its simplicity, we use BYOL to obtain concise embeddings from our demonstrated data.

\subsection{Non-Parametric Imitation Learning} \label{sec:il_background}
In our framework for imitation learning we have access to expert demonstrations in the form of $\mathcal{D}^{E} \equiv \{(o^{E}_{t'},s^{E}_{t'},a^{E}_{t'})\}$, where $o^E_{t'}$ represents the sensory observation at time ${t'}$, $s^E_{t'}$ represents the robot state, and $a^E_{t'}$ denotes the robot action taken. Note that $s^E_{t'}$ does not contain information about the object that is being manipulated. Hence, object information needs to be inferred from observations $o^{E}_{t'}$. Given these demonstrations, we would like to learn a policy $\pi(a_t | o_t)$ that follows the expert behavior $\mathcal{D}^E$. While there are several strategies for optimizing $\pi$, we resort to non-parametric approaches given their superior performance in low-data regimes~\cite{pari2021surprising,arunachalam2022dexterous}.

Our non-parametric control framework follows VINN~\cite{pari2021surprising}, where given the observations $o^E$ from the expert demonstration dataset $\mathcal{D}^E$, a BYOL encoder $f_{\theta}$ is trained. Next, the observations in the dataset are all converted to embeddings, i.e. $\{o^E_{t'}\} \xrightarrow[]{f_{\theta}} \{z^E_{t'}\}$. During run time, when the robot receives an observation $o_t$ it is embedded to $z_t$. Then the Nearest-Neighbor (NN) example in $\{(z^E_{t'},s^E_{t'},a^E_{t'})\}$ is selected to be imitated. We denote this NN example as $\{(z^E_{t^*},s^E_{t^*},a^E_{t^*})\}$. Given a small dataset $\mathcal{D}^E$, which is often the case in robotic applications, this NN-based imitation learning provides effective learning compared to parametric approaches such as BC.

\section{\method{}}

\begin{figure*}[t!]
    \centering
    \includegraphics[width=\linewidth]{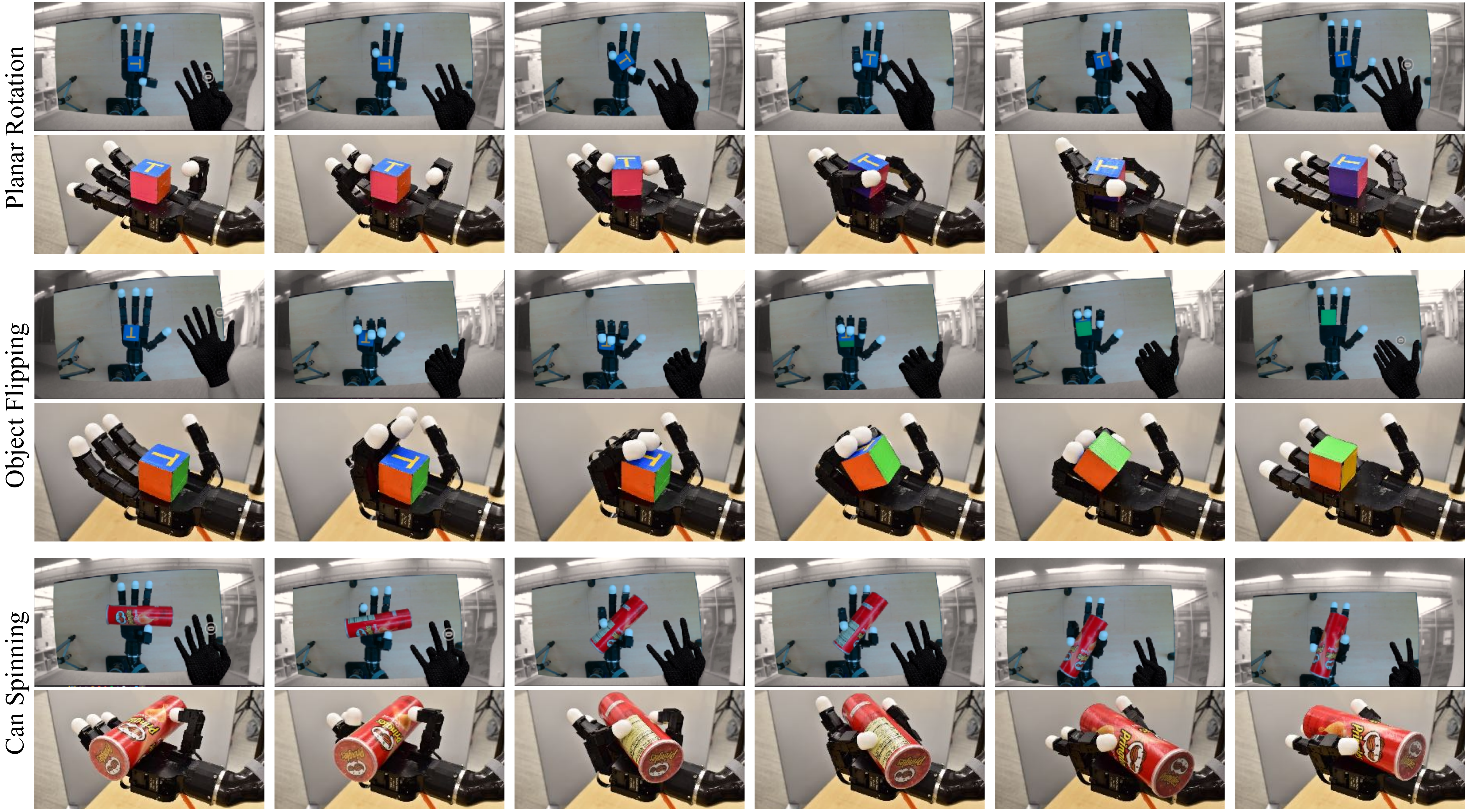}
    \caption{Demonstration collection process for three of our tasks. For each task, the first row shows the user's perspective inside the VR Headset and the second row shows the corresponding robot hand configuration.}
    \label{fig:demos}
    \vspace{-0.1in}
\end{figure*}

As seen in Fig.~\ref{fig:intro}, \method{} operates in two phases. In the first phase, a human teacher uses a Virtual Reality (VR) headset to provide demonstrations to a robot. This phase consists of creating a virtual world for teaching, estimating hand poses from the teacher, retargeting the teacher's hand pose to the robot's hand and finally controlling the robot hand. After a handful of demonstrations are collected in phase one, the second phase of \method{} learns visual policies to solve the demonstrated tasks. In this section, we will describe each sub-component in detail.
% \SA{David's comment: This para is not related to the sub-sections}

\subsection{Placing an Operator in a Virtual World}
We use the Meta Quest 2 VR headset to place human teachers in a virtual world. The headset surrounds the human in a virtual environment at a resolution of $1832\times 1920$ and a refresh rate of $72$ Hz. The base version of this headset is affordable at \$$399$ and is relatively light at $503$g. These features allow for comfortable operation by the teacher. Importantly, the API interface of the Quest 2 allows for creating custom mixed reality worlds that visualizes the robotic system along with diagnostic panels in VR. Examples of virtual scenes are depicted in Fig.~\ref{fig:aug2real} and Fig.~\ref{fig:demos}.

\subsection{Hand Pose Estimation with VR Headsets}
In contrast to prior work on dexterous teleoperation, using VR headsets provides three benefits with regard to hand pose estimation of the human teacher. First, since the Quest 2 uses 4 monochrome cameras, its hand-pose estimator~\cite{megatrack} is significantly more robust compared to single camera estimators~\cite{zhang2020mediapipe}. Second, since the cameras are internally calibrated, they do not require specialized calibration routines that are needed in prior multi-camera teleoperation frameworks~\cite{DexPilot}. Third, since the hand pose estimator is integrated into the device, it can stream real-time poses at 72Hz. As noted in prior work~\cite{DexPilot,arunachalam2022dexterous}, a significant challenge in dexterous teleoperation is obtaining hand poses at both high accuracy and a high frequency. \method{} significantly simplifies this problem by using commercial-grade VR headsets.

\begin{figure*}[t!]
    \centering
    \includegraphics[width=\linewidth]{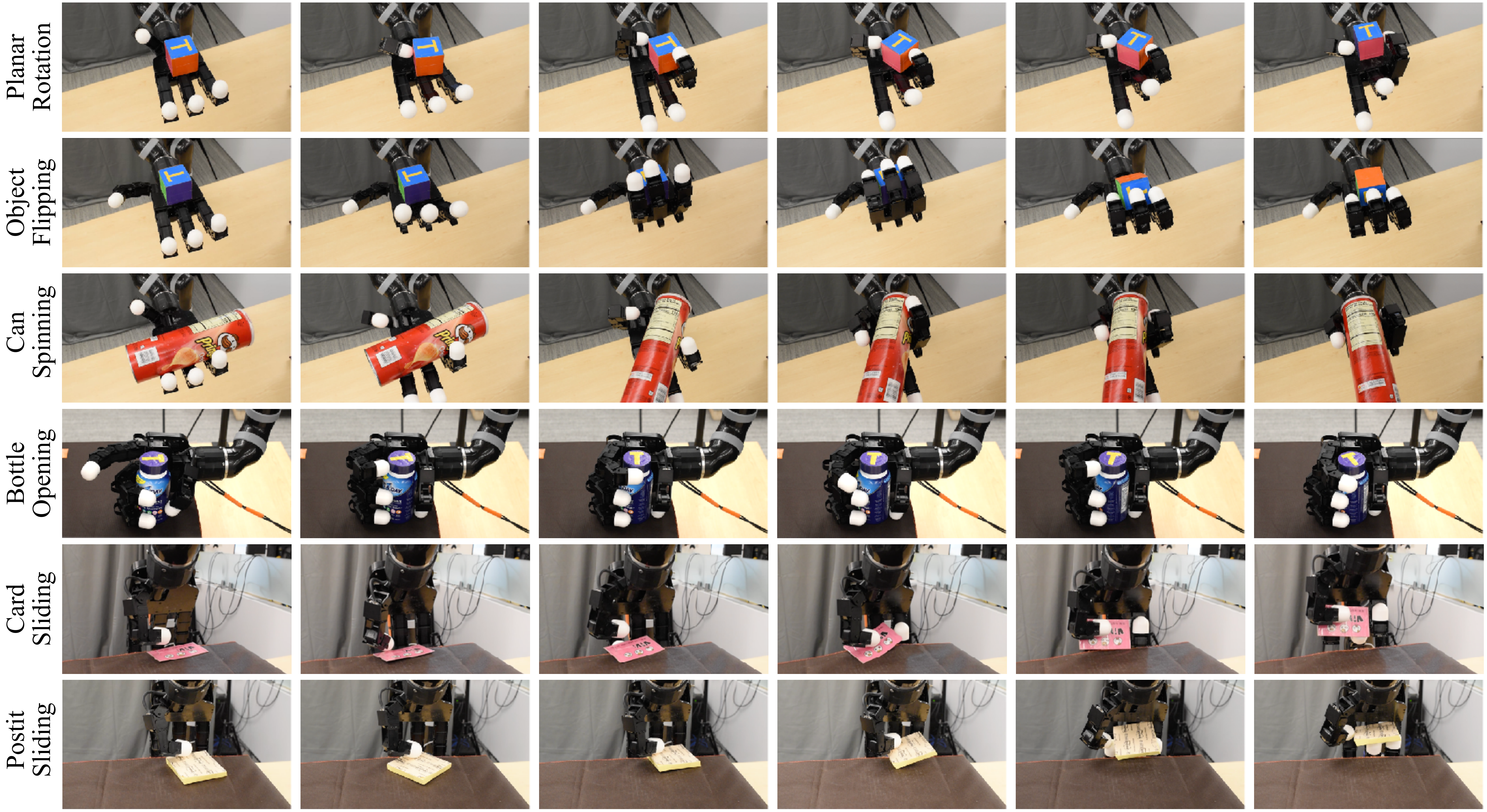}
    \caption{Successful rollouts of visual policies trained through \method{} on our six dexterous tasks.}
    \label{fig:policy}
    \vspace{-0.1in}
\end{figure*}

\subsection{Human to Robot Hand Pose Retargeting}
Once we have extracted the teacher's hand pose from VR, we will need to retarget it to the robotic hand. This is done by first computing the individual hand joint angles in the teacher's hand. Given these joint angles, a straightforward method of retargeting is to directly command the robot's joints to the corresponding angle. In practice, this works well for all fingers except the thumb. The thumb presents a unique challenge to our Allegro robot hand since its morphology does not match a human's hand. To address this, we map the spatial coordinates of the teacher's thumb fingertip to the robot's thumb fingertip. The joint angles of the thumb are then computed through an inverse kinematics solver. Since the Allegro hand does not have a pinky finger, we ignore the teacher's pinky joints. 

The overall pose retargeting procedure does not require any calibration or user-specific tuning to collect demonstrations. However, we find that thumb retargeting can be improved by finding user-specific maps from their thumb to the robot's thumb. This entire procedure is computationally inexpensive and can stream desired robot hand poses at 60 Hz.

\subsection{Robot Hand Control}
% \LP{How does the robot achieve the desired hand pose from human? Add all details pertaining to teleoperation and obtaining data here.}

Our Allegro Hand is controlled asynchronously over a ROS~\cite{ros} communication framework. Given desired robot joint positions that were computed from the retargeting procedure, we use a PD controller to output desired torques at 300Hz. To reduce steady-state error, we use a gravity compensation module to compute offset torques. On latency tests, we find that when the VR headset is on the same local network as the robot hand, we achieve latency under 100 milliseconds. Having a low error and latency is crucial for \method{} since it allows for intuitive teleoperation of the robot hand by the human teacher.

As the human teacher controls the robot hand, they can see the robot change in real time (60Hz). This allows the teacher to correct execution errors in the robot. During the teaching process, we record observational data from three RGBD cameras and the action information of the robot at 5Hz. We had to reduce the recording frequency due to the large data footprint and associated bandwidth required from recording multiple cameras.

\subsection{Imitation Learning with \method{} Data}
Once data is collected in the first phase of \method{}, we now proceed to the second phase, where visual policies are trained on top of this data. We employ the Imitation with Nearest Neighbors (INN) algorithm for learning. Background details of INN is present in Section~\ref{sec:il_background}. In prior work, INN was shown to produce state-based dexterous policies on the Allegro hand~\cite{arunachalam2022dexterous}. \method{} takes this a few steps further and demonstrates that these visual policies can generalize to novel objects in a variety of dexterous manipulation tasks.

To select the learning algorithm for obtaining low-dimensional embeddings (see Section~\ref{sec:ssl_background}), we experiment with several state-of-the-art self-supervised learning algorithms~\cite{grill2020bootstrap,chen2020simple,chen2021empirical,bardes2021vicreg} and find that BYOL~\cite{grill2020bootstrap} provides the best nearest neighbour results. Hence we select BYOL as our base self-supervised learning method. Once BYOL is trained on the collected demonstrations, we can run dexterous manipulation policies on the robot by performing nearest neighbor action retrieval (see Section~\ref{sec:il_background}) to get the closest example in the trainset $(z^E_{t^*},a^E_{t^*})$. To account for the slower rate of demonstration collection, we set the action to the difference between the succeeding state to the closest neighbor and the current state, i.e. $a_t = \mathcal(s^E_{t^*+k}) - (s^E_{t})$. Here $k$ represents the number of states skipped during our recording of demonstrations. Note that directly commanding $a^E_{t^*}$ would fail due to our asynchronous data storage framework.

%%%%%%%%%%%%%%%%%%%%%%%%%%%%%%%%%%%%%%%%%%%%%%%%%%%%%%%%%%%%%%%%%%%%%%%%%%%%%%%%
\section{Experimental Evaluation}

Our experiments and tasks are designed to answer the following questions:
\begin{itemize}
    \item How long does it take \method{} to collect demonstrations? 
    \item How successful are policies trained by \method{}?
    \item How general are the skills learned by \method{}?
    \item How many demonstrations from \method{} are required to successfully solve dexterous tasks?
\end{itemize}

\subsection{Dexterous Manipulation Tasks}
We study six dexterous manipulation tasks that require contact-rich, multi-fingered control for successful completion. Details of these tasks are described below.

\begin{enumerate}
    \item \textit{Planar Rotation:} Given an object placed at a random position on the palm of the robot hand, the goal is to rotate the object in the counter-clockwise direction along the palm normal vector. Solving this task requires the robot to make multi-fingered contacts to both rotate and correct for deviations of the object from the center of the hand. The task is considered a success if the robot is able to rotate the object by $90^{\circ}$ under a minute. 
    
    \item \textit{Object Flipping:} Given an object placed at a random position on the palm of the robot, the goal is to flip the object in the hand's direction. Solving this task requires the robot to make multi-fingered contacts for grasping the top face of the object and to correct the object when it deviates from the center of the hand. The task is considered a success if the robot is able to flip the object by $90^{\circ}$ within a minute.

    \item \textit{Can Spinning:} Given a large can placed horizontally on the palm of the robot hand, the task is to spin the can in the counter-clockwise direction along the palm normal vector. Solving this task requires the robot to make synchronized multi-fingered contacts to apply controlled torques on the curved sides of the can. The task is considered a success if the robot is able to spin the can by $90^{\circ}$ under 30 seconds. 
    
    \item \textit{Bottle Opening}: Given a bottle on a desk, the task is to single-handedly grab the bottle and turn its lid open using the index finger. Solving this task requires the robot to use the index finger to grip the bottle cap and turn it while using the non-index fingers to hold the bottle firmly. The task is considered a success when the robot rotates the bottle cap by $360^{\circ}$ under 180 seconds.

    \item \textit{Card Sliding:} Given a card on a desk in front of the hand, the task is to grab the card by sliding it off the table and picking it up. To solve the task the robot requires to use the thumb finger to slide the card to the edge of the desk and the other non-thumb fingers to grab the card from the edge. The task is considered a success when the robot is able to lift the card off and stably grasp it from the table under 120 seconds. 
    
    \item \textit{PostIt Note Sliding}: This task is similar to card sliding, but instead of a card we use a thicker post-it note pad as the object to pick from the desk. To solve the task, the robot needs to apply a firmer torque on the post-it note pad using the thumb finger since the object is heavier. The task is considered a success when the robot is able to lift the card off and stably grasp it from the table under 120 seconds.
    
\end{enumerate}

For the \textit{Planar Rotation} task we collect $120$ expert demonstrations, while for all the other tasks we collect $30$ demonstrations. Additional demonstrations for \textit{Planar Rotation} are collected to account for the relative difficulty of this task and experimentation with different dataset sizes.

\subsection{How long does it take to collect demonstrations?}

\begin{figure*}[ht!]
    \centering
    \includegraphics[width=\linewidth]{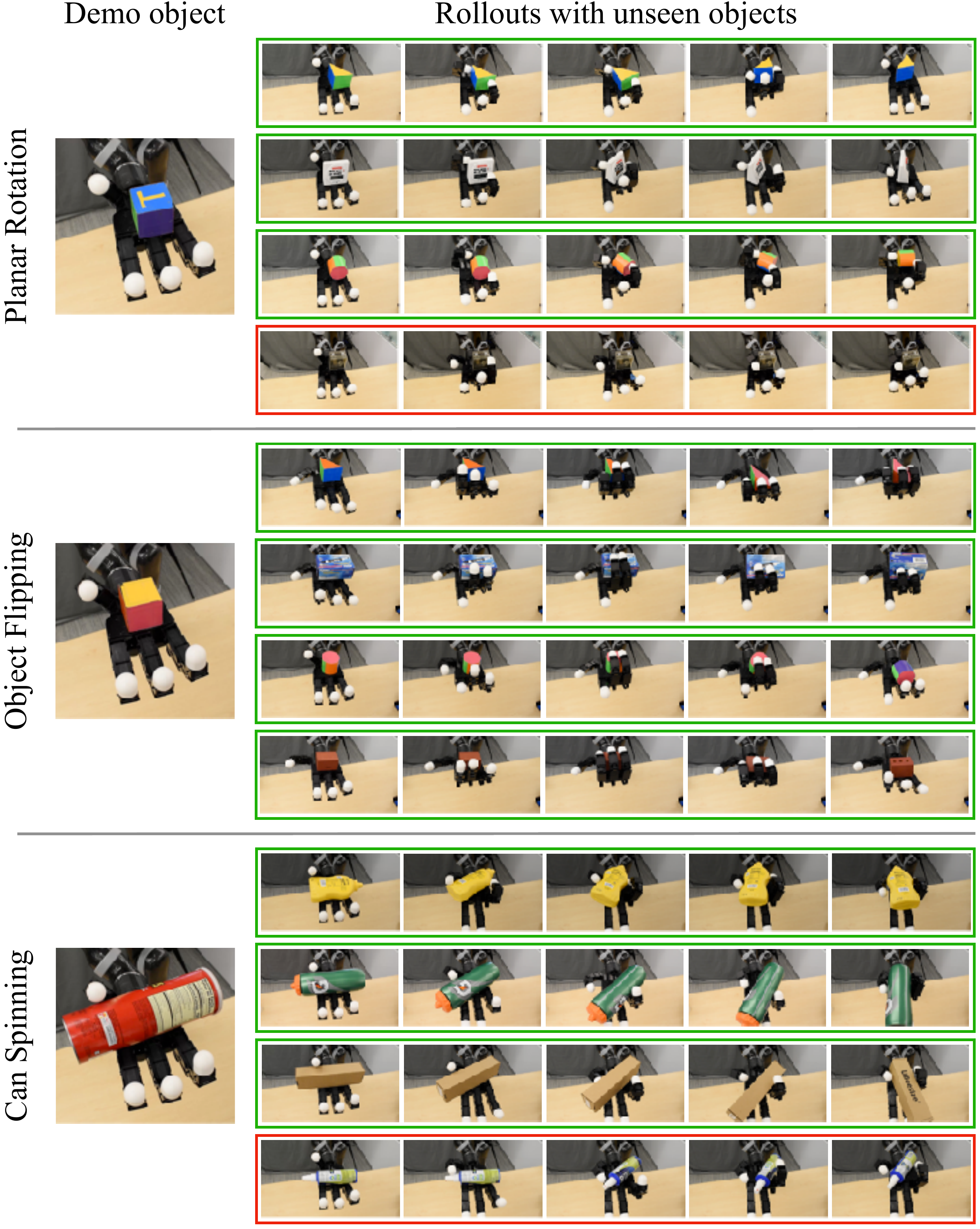}
    \caption{On the left, we depict the object present in demonstration data. On the right, we depict the rollouts produced by running our policies on objects that were not present in demonstration collection. Green boxes denote a successful rollout, while red boxes denote a failure. We see that policies learned by \method{} are fairly robust to visually diverse novel objects without object-specific training.}
    \label{fig:object_generalization}
    \vspace{-0.1in}
\end{figure*}

The closest dexterous teleoperation work to ours that uses commodity sensors is single-image teleoperation (e.g. DIME~\cite{arunachalam2022dexterous}), where hand poses are detected via RGB images to get robot joint angles. 
Despite its simplicity, such single-image pose estimation~\cite{zhang2020mediapipe} suffers from hand occlusions, which results in poor teleoperation performance on challenging manipulation tasks~\cite{DexPilot}. 
In Table~\ref{tab:demo_collection_timings} we show that \method{} can collect successful demonstrations $1.8\times$ faster compared to DIME. For $3/6$ tasks that require precise 3D movements, we find that single-image teleoperation is insufficient to collect even a single demonstration.

To demonstrate the versatility of \method{} we ask five untrained users to collect demonstrations for each of our tasks. Unlike prior work~\cite{arunachalam2022dexterous, DexPilot}, no user-specific calibration was done for this evaluation. In Table~\ref{tab:demo_collection_timings}, we see that these users are successfully able to solve $4/6$ tasks on their first try, failing only on more difficult sliding tasks. We also find that training on this system is quite important as it yields a nearly $2.6\times$ speedup in demonstration collection.

\subsection{How successful are policies trained by \method{}?}

We examine the performance of various imitation learning policies on all dexterous tasks. The imitation learning algorithms include Behavior Cloning (BC)~\cite{pomerleau1989alvinn}, Behavior Cloning from pretrained representations (BC-Rep)~\cite{young2021playful} and VINN~\cite{pari2021surprising}. 
Table~\ref{tab:rob_results} shows the success rates of each task with different policies. We find that VINN outperforms both Behavior Cloning algorithms on all tasks. This is in line with prior work~\cite{pari2021surprising,arunachalam2022dexterous} and showcases the effectiveness of non-parametric imitation with few demonstrations. However, we find that for the two tasks that involve sliding and picking, the performance of VINN is quite low at $30\%$. We believe this is due to our robot's inability to sense touch, which limits our vision-only model from performing precise actions. 

\begin{table}[t!]
\caption{Average time taken in seconds to collect a single demonstration on our Allegro hand using \method{} and DIME~\cite{arunachalam2022dexterous}}
\centering
\begin{tabular}{@{}ccccc@{}}
\toprule
\multirow{2}{*}{Task} & \multicolumn{2}{c}{DIME} & \multicolumn{2}{c}{\method{}} \\ \cmidrule(l){2-5} 
                      & Expert     & New User    & Expert                & New User               \\ \midrule
Planar Rotation       & 60         & 150         & 30                    & 125                    \\
Object Flipping       & 6          & 21          & 5                     & 6                      \\
Can Spinning          & 15         & 76          & 10                    & 68                     \\
Bottle Opening        & N/A        & N/A         & 30                    & 48                     \\
Card Sliding          & N/A        & N/A         & 150                   & N/A                    \\
PostIt Note Sliding   & N/A        & N/A         & 120                   & N/A                    \\ \bottomrule
\end{tabular}
\label{tab:demo_collection_timings}
\vspace{-0.2in}
\end{table}

\begin{table}[h!]
\caption{Success rates on our Allegro hand using \method{}}
\centering
\begin{tabular}{@{}ccccc@{}}
\toprule
Task                & BC   & BC - Rep & VINN  & \begin{tabular}[c]{@{}c@{}}VINN\\ (New Objects)\end{tabular} \\ \midrule
Planar Rotation     & 0/10 & 0/10     & 10/10 & 35/50                                                        \\
Object Flipping     & 0/10 & 0/10     & 9/10  & 50/50                                                        \\
Can Spinning        & 0/10 & 0/10     & 10/10 & 45/50                                                        \\
Bottle Opening      & 0/10 & 0/10     & 10/10 & N/A                                                          \\
Card Sliding        & 0/10 & 0/10     & 3/10  & N/A                                                          \\
PostIt Note Sliding & 0/10 & 0/10     & 3/10  & N/A                                                          \\ \bottomrule
\end{tabular}
\label{tab:rob_results}
\vspace{-0.2in}
\end{table}

\subsection{How general are the policies learned by \method{}?}

To understand the generalization capabilities of our models, we analyse the quality of the embeddings we get for a given visual input. Interestingly, we observed that the Planar Rotation policy's encoder was able to generalize to the other two in-hand manipulation tasks. We reason that this encoder was able to exhibit this behavior since it was trained on abundant Planar Rotation data, where the object used while collecting demonstrations had different colors on it's each face and was placed on various locations. 

Since our policies are vision-based and do not require explicitly estimating the states of objects, they are compatible with objects not seen in training. We evaluate our in-hand manipulation policies which were trained for performing Planar Rotation, Object Flipping, and Can Spinning tasks on 10 visually and geometrically distinct objects each, with 5 rollouts for every object in different initial positions. Results for this experiment are in Table~\ref{tab:rob_results} and are visualized in Fig.~\ref{fig:object_generalization}. Surprisingly, for all three in-hand manipulation tasks, we find high success rates without any additional demonstration collection or training. We observed that the Planar Rotation policy was able to generalize on 7 out of 10 objects, whereas the Object Flipping and Can Spinning policies were able to succeed at performing the task on 10 and 9 unseen objects respectively. We believe that the policy fails to generalize on some objects because of their visual features (object color and shape) being very different from that of the object in the demonstrations. This means that although we collect demonstrations from \method{} on a single object, the learned policies can generalize in a zero-shot manner.

\subsection{How many demonstrations are needed to solve our tasks? }

\begin{figure}[t!]
    \centering
    \includegraphics[width=\linewidth]{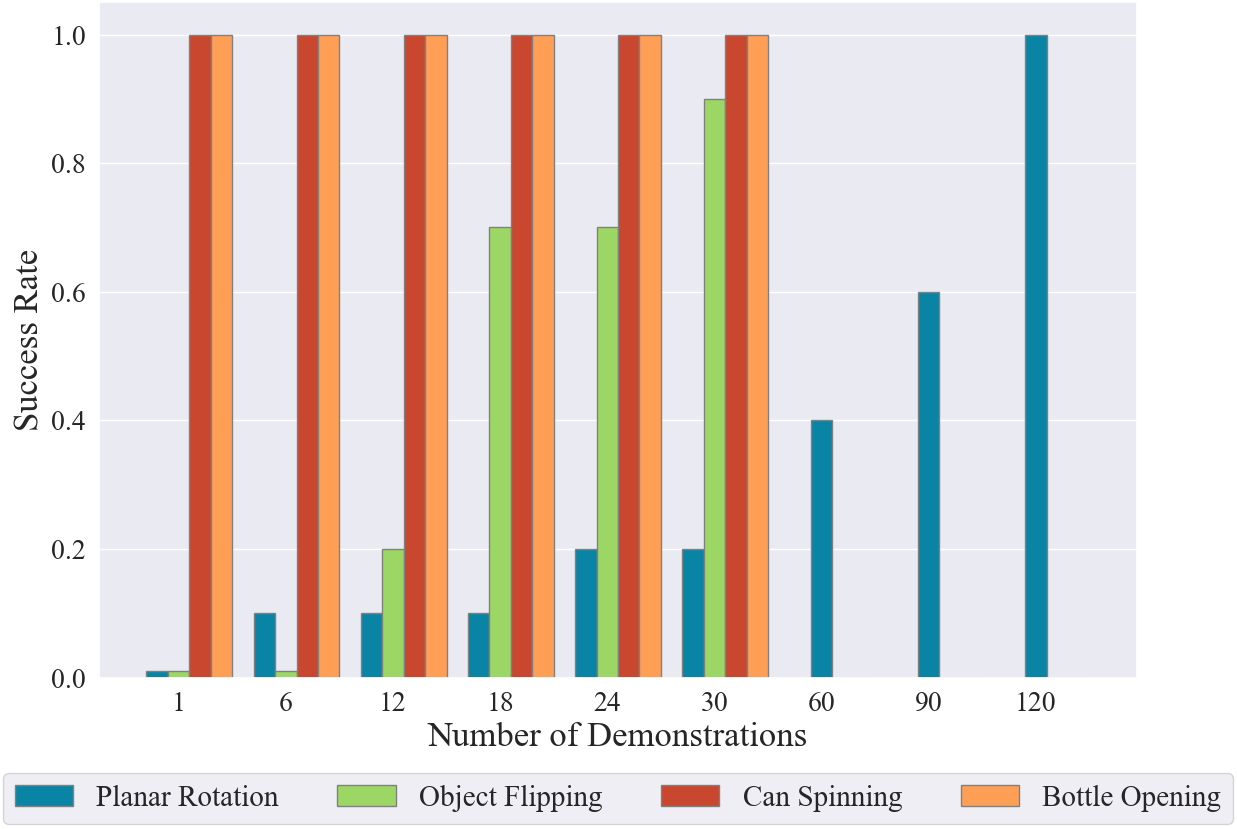}
    \caption{Effect of varying demonstration data with \method{}.}
    \label{fig:limited_data_graph}
    \vspace{-0.2in}
\end{figure}

In Fig.~\ref{fig:limited_data_graph} we visualize the performance on four of our tasks across different dataset sizes. To decouple the effects of representation learning with action prediction, we use the same encoder (trained on all task data) for different dataset splits. We find that for Can Spinning and Bottle Opening, a single demonstration is sufficient to achieve high performance, while for Planar Rotation and Object Flipping we see steady gains in performance as we increase the amount of demonstration data.

%%%%%%%%%%%%%%%%%%%%%%%%%%%%%%%%%%%%%%%%%%%%%%%%%%%%%%%%%%%%%%%%%%%%%%%%%%%%%%%%

\section{Limitations and Discussion}

We have presented \method{}, a framework that takes some of the first steps towards immersive teaching of dexterous robots through VR. There are currently two limitations of this work. First, we find that for the harder manipulation tasks, such as \textit{sliding a card} our learned policies achieve poor performance. Integrating tactile sensing to \method{} could remedy this issue. Second, our retargeting procedure only applies to robots that can map to human joints. This limits its applicability to robots with different morphologies~(e.g. aerial robots, quadrupeds, etc.). Future research on UX design and retargeting mechanisms can enable mapping VR control to more complex end-effectors.

%%%%%%%%%%%%%%%%%%%%%%%%%%%%%%%%%%%%%%%%%%%%%%%%%%%%%%%%%%%%%%%%%%%%%%%%%%%%%%%%
\section{Acknowledgements}

We thank Ankur Handa, Ilija Radosavovic, Josh Merel, David Brandfonbrener, Ben Evans, Mahi Shaffiulah, Jeff Cui, Jyo Pari and Siddhanth Haldar for feedback and discussions. This work was supported by a Honda award and ONR award N000142112758.

%%%%%%%%%%%%%%%%%%%%%%%%%%%%%%%%%%%%%%%%%%%%%%%%%%%%%%%%%%%%%%%%%%%%%%%%%%%%%%%%

\bibliographystyle{IEEEtran}
\small
\bibliography{IEEEexample}

\begin{thebibliography}{10}
\providecommand{\url}[1]{#1}
\csname url@rmstyle\endcsname
\providecommand{\newblock}{\relax}
\providecommand{\bibinfo}[2]{#2}
\providecommand\BIBentrySTDinterwordspacing{\spaceskip=0pt\relax}
\providecommand\BIBentryALTinterwordstretchfactor{4}
\providecommand\BIBentryALTinterwordspacing{\spaceskip=\fontdimen2\font plus
\BIBentryALTinterwordstretchfactor\fontdimen3\font minus
  \fontdimen4\font\relax}
\providecommand\BIBforeignlanguage[2]{{%
\expandafter\ifx\csname l@#1\endcsname\relax
\typeout{** WARNING: IEEEtran.bst: No hyphenation pattern has been}%
\typeout{** loaded for the language `#1'. Using the pattern for}%
\typeout{** the default language instead.}%
\else
\language=\csname l@#1\endcsname
\fi
#2}}

\bibitem{chi2022iterative}
C.~Chi, B.~Burchfiel, E.~Cousineau, S.~Feng, and S.~Song, ``Iterative residual
  policy: for goal-conditioned dynamic manipulation of deformable objects,''
  \emph{arXiv preprint arXiv:2203.00663}, 2022.

\bibitem{tekden2020belief}
A.~E. Tekden, A.~Erdem, E.~Erdem, M.~Imre, M.~Y. Seker, and E.~Ugur, ``Belief
  regulated dual propagation nets for learning action effects on groups of
  articulated objects,'' in \emph{2020 IEEE International Conference on
  Robotics and Automation (ICRA)}, 2020, pp. 10\,556--10\,562.

\bibitem{gangapurwala2022rloc}
S.~Gangapurwala, M.~Geisert, R.~Orsolino, M.~Fallon, and I.~Havoutis, ``Rloc:
  Terrain-aware legged locomotion using reinforcement learning and optimal
  control,'' \emph{IEEE Transactions on Robotics}, 2022.

\bibitem{ma2022combining}
Y.~Ma, F.~Farshidian, T.~Miki, J.~Lee, and M.~Hutter, ``Combining
  learning-based locomotion policy with model-based manipulation for legged
  mobile manipulators,'' \emph{IEEE Robotics and Automation Letters}, vol.~7,
  no.~2, pp. 2377--2384, 2022.

\bibitem{smith2022walk}
L.~Smith, I.~Kostrikov, and S.~Levine, ``A walk in the park: Learning to walk
  in 20 minutes with model-free reinforcement learning,'' \emph{arXiv preprint
  arXiv:2208.07860}, 2022.

\bibitem{zhang2016learning}
T.~Zhang, G.~Kahn, S.~Levine, and P.~Abbeel, ``Learning deep control policies
  for autonomous aerial vehicles with mpc-guided policy search,'' in \emph{2016
  IEEE international conference on robotics and automation (ICRA)}.\hskip 1em
  plus 0.5em minus 0.4em\relax IEEE, 2016, pp. 528--535.

\bibitem{gandhi2017learning}
D.~Gandhi, L.~Pinto, and A.~Gupta, ``Learning to fly by crashing,'' in
  \emph{2017 IEEE/RSJ International Conference on Intelligent Robots and
  Systems (IROS)}.\hskip 1em plus 0.5em minus 0.4em\relax IEEE, 2017, pp.
  3948--3955.

\bibitem{hwangbo2017control}
J.~Hwangbo, I.~Sa, R.~Siegwart, and M.~Hutter, ``Control of a quadrotor with
  reinforcement learning,'' \emph{IEEE Robotics and Automation Letters},
  vol.~2, no.~4, pp. 2096--2103, 2017.

\bibitem{berner2019dota}
C.~Berner, G.~Brockman, B.~Chan, V.~Cheung, P.~Debiak, C.~Dennison, D.~Farhi,
  Q.~Fischer, S.~Hashme, C.~Hesse, \emph{et~al.}, ``Dota 2 with large scale
  deep reinforcement learning,'' \emph{arXiv preprint arXiv:1912.06680}, 2019.

\bibitem{silver2016mastering}
D.~Silver, A.~Huang, C.~J. Maddison, A.~Guez, L.~Sifre, G.~Van Den~Driessche,
  J.~Schrittwieser, I.~Antonoglou, V.~Panneershelvam, M.~Lanctot,
  \emph{et~al.}, ``Mastering the game of go with deep neural networks and tree
  search,'' \emph{nature}, vol. 529, no. 7587, pp. 484--489, 2016.

\bibitem{guo2018long}
J.~Guo, S.~Lu, H.~Cai, W.~Zhang, Y.~Yu, and J.~Wang, ``Long text generation via
  adversarial training with leaked information,'' in \emph{Proceedings of the
  AAAI conference on artificial intelligence}, vol.~32, no.~1, 2018.

\bibitem{huang2021sarg}
M.~Huang, F.~Li, W.~Zou, and W.~Zhang, ``Sarg: A novel semi autoregressive
  generator for multi-turn incomplete utterance restoration,'' in
  \emph{Proceedings of the AAAI Conference on Artificial Intelligence},
  vol.~35, no.~14, 2021, pp. 13\,055--13\,063.

\bibitem{dosovitskiy2020image}
A.~Dosovitskiy, L.~Beyer, A.~Kolesnikov, D.~Weissenborn, X.~Zhai,
  T.~Unterthiner, M.~Dehghani, M.~Minderer, G.~Heigold, S.~Gelly,
  \emph{et~al.}, ``An image is worth 16x16 words: Transformers for image
  recognition at scale,'' \emph{arXiv preprint arXiv:2010.11929}, 2020.

\bibitem{gidaris2018dynamic}
S.~Gidaris and N.~Komodakis, ``Dynamic few-shot visual learning without
  forgetting,'' in \emph{Proceedings of the IEEE conference on computer vision
  and pattern recognition}, 2018, pp. 4367--4375.

\bibitem{pinto2016supersizing}
L.~Pinto and A.~Gupta, ``Supersizing self-supervision: Learning to grasp from
  50k tries and 700 robot hours,'' \emph{ICRA}, 2016.

\bibitem{levine2016learning}
S.~Levine, P.~Pastor, A.~Krizhevsky, and D.~Quillen, ``Learning hand-eye
  coordination for robotic grasping with deep learning and large-scale data
  collection,'' \emph{ISER}, 2016.

\bibitem{singh2019end}
A.~Singh, L.~Yang, K.~Hartikainen, C.~Finn, and S.~Levine, ``End-to-end robotic
  reinforcement learning without reward engineering,'' \emph{arXiv preprint
  arXiv:1904.07854}, 2019.

\bibitem{vecerik2019practical}
M.~Vecerik, O.~Sushkov, D.~Barker, T.~Roth{\"o}rl, T.~Hester, and J.~Scholz,
  ``A practical approach to insertion with variable socket position using deep
  reinforcement learning,'' in \emph{2019 international conference on robotics
  and automation (ICRA)}.\hskip 1em plus 0.5em minus 0.4em\relax IEEE, 2019,
  pp. 754--760.

\bibitem{dulac2020empirical}
G.~Dulac-Arnold, N.~Levine, D.~J. Mankowitz, J.~Li, C.~Paduraru, S.~Gowal, and
  T.~Hester, ``An empirical investigation of the challenges of real-world
  reinforcement learning,'' \emph{arXiv preprint arXiv:2003.11881}, 2020.

\bibitem{Openai2018}
OpenAI, M.~Andrychowicz, B.~Baker, M.~Chociej, R.~Józefowicz, B.~McGrew,
  J.~Pachocki, A.~Petron, M.~Plappert, G.~Powell, A.~Ray, J.~Schneider,
  S.~Sidor, J.~Tobin, P.~Welinder, L.~Weng, and W.~Zaremba, ``Learning
  dexterous in-hand manipulation,'' \emph{arXiv}, 2018.

\bibitem{Openai2019}
OpenAI, I.~Akkaya, M.~Andrychowicz, M.~Chociej, M.~Litwin, B.~McGrew,
  A.~Petron, A.~Paino, M.~Plappert, G.~Powell, R.~Ribas, J.~Schneider,
  N.~Tezak, J.~Tworek, P.~Welinder, L.~Weng, Q.~Yuan, W.~Zaremba, and L.~Zhang,
  ``Solving rubik's cube with a robot hand,'' \emph{arXiv}, 2019.

\bibitem{tobin2017domain}
J.~Tobin, R.~Fong, A.~Ray, J.~Schneider, W.~Zaremba, and P.~Abbeel, ``Domain
  randomization for transferring deep neural networks from simulation to the
  real world,'' in \emph{IROS}, 2017.

\bibitem{jakobi1995sim}
N.~Jakobi, P.~Husbands, and I.~Harvey, ``Noise and the reality gap: The use of
  simulation in evolutionary robotics,'' in \emph{Advances in Artificial Life},
  F.~Mor{\'a}n, A.~Moreno, J.~J. Merelo, and P.~Chac{\'o}n, Eds.\hskip 1em plus
  0.5em minus 0.4em\relax Berlin, Heidelberg: Springer Berlin Heidelberg, 1995,
  pp. 704--720.

\bibitem{pomerleau1989alvinn}
D.~A. Pomerleau, ``Alvinn: An autonomous land vehicle in a neural network,'' in
  \emph{NeurIPS}, 1989, pp. 305--313.

\bibitem{abbeel2004apprenticeship}
P.~Abbeel and A.~Y. Ng, ``Apprenticeship learning via inverse reinforcement
  learning,'' in \emph{ICML}, 2004, p.~1.

\bibitem{rajeswaran2017learning}
A.~Rajeswaran, V.~Kumar, A.~Gupta, G.~Vezzani, J.~Schulman, E.~Todorov, and
  S.~Levine, ``Learning complex dexterous manipulation with deep reinforcement
  learning and demonstrations,'' \emph{RSS}, 2018.

\bibitem{DexPilot}
A.~Handa, K.~Van~Wyk, W.~Yang, J.~Liang, Y.-W. Chao, Q.~Wan, S.~Birchfield,
  N.~Ratliff, and D.~Fox, ``Dexpilot: Vision-based teleoperation of dexterous
  robotic hand-arm system,'' in \emph{2020 IEEE International Conference on
  Robotics and Automation (ICRA)}, 2020, pp. 9164--9170.

\bibitem{arunachalam2022dexterous}
S.~P. Arunachalam, S.~Silwal, B.~Evans, and L.~Pinto, ``Dexterous imitation
  made easy: A learning-based framework for efficient dexterous manipulation,''
  \emph{arXiv preprint arXiv:2203.13251}, 2022.

\bibitem{glovereview}
M.~Caeiro-Rodríguez, I.~Otero-González, F.~A. Mikic-Fonte, and
  M.~Llamas-Nistal, ``A systematic review of commercial smart gloves: Current
  status and applications,'' \emph{Sensors}, 2021.

\bibitem{Kaelbling1996}
L.~P. Kaelbling, M.~L. Littman, and A.~W. Moore, ``Reinforcement learning: A
  survey,'' \emph{JAIR}, 1996.

\bibitem{lillicrap2015continuous}
T.~P. Lillicrap, J.~J. Hunt, A.~Pritzel, N.~Heess, T.~Erez, Y.~Tassa,
  D.~Silver, and D.~Wierstra, ``Continuous control with deep reinforcement
  learning,'' \emph{arXiv preprint}, 2015.

\bibitem{yarats2021mastering}
D.~Yarats, R.~Fergus, A.~Lazaric, and L.~Pinto, ``Mastering visual continuous
  control: Improved data-augmented reinforcement learning,'' \emph{arXiv
  preprint arXiv:2107.09645}, 2021.

\bibitem{sadeghi2016cad2rl}
F.~Sadeghi and S.~Levine, ``Cad2rl: Real single-image flight without a single
  real image,'' \emph{arXiv preprint arXiv:1611.04201}, 2016.

\bibitem{pinto2017asymmetric}
L.~Pinto, M.~Andrychowicz, P.~Welinder, W.~Zaremba, and P.~Abbeel, ``Asymmetric
  actor critic for image-based robot learning,'' \emph{RSS}, 2018.

\bibitem{wu2019learning}
Y.~Wu, W.~Yan, T.~Kurutach, L.~Pinto, and P.~Abbeel, ``Learning to manipulate
  deformable objects without demonstrations,'' \emph{arXiv preprint}, 2019.

\bibitem{rao2020rl}
K.~Rao, C.~Harris, A.~Irpan, S.~Levine, J.~Ibarz, and M.~Khansari,
  ``Rl-cyclegan: Reinforcement learning aware simulation-to-real,'' in
  \emph{Proceedings of the IEEE/CVF Conference on Computer Vision and Pattern
  Recognition}, 2020, pp. 11\,157--11\,166.

\bibitem{qin2022one}
Y.~Qin, H.~Su, and X.~Wang, ``From one hand to multiple hands: Imitation
  learning for dexterous manipulation from single-camera teleoperation,''
  \emph{arXiv preprint arXiv:2204.12490}, 2022.

\bibitem{florence2021implicit}
P.~Florence, C.~Lynch, A.~Zeng, O.~Ramirez, A.~Wahid, L.~Downs, A.~Wong,
  J.~Lee, I.~Mordatch, and J.~Tompson, ``Implicit behavioral cloning,'' 2021.

\bibitem{bojarski2016end}
M.~Bojarski, D.~Del~Testa, D.~Dworakowski, B.~Firner, B.~Flepp, P.~Goyal, L.~D.
  Jackel, M.~Monfort, U.~Muller, J.~Zhang, \emph{et~al.}, ``End to end learning
  for self-driving cars,'' \emph{arXiv:1604.07316}, 2016.

\bibitem{young2020visual}
S.~Young, D.~Gandhi, S.~Tulsiani, A.~Gupta, P.~Abbeel, and L.~Pinto, ``Visual
  imitation made easy,'' 2020.

\bibitem{rob_lwr}
S.~Schaal and C.~Atkeson, ``Robot juggling: implementation of memory-based
  learning,'' \emph{IEEE CSM}, 1994.

\bibitem{pari2021surprising}
J.~Pari, N.~M. Shafiullah, S.~P. Arunachalam, and L.~Pinto, ``The surprising
  effectiveness of representation learning for visual imitation,'' 2021.

\bibitem{IRL_gt}
U.~Syed and R.~E. Schapire, ``A game-theoretic approach to apprenticeship
  learning,'' in \emph{NeruIPS}, J.~Platt, D.~Koller, Y.~Singer, and S.~Roweis,
  Eds., 2007.

\bibitem{haldar2022watch}
S.~Haldar, V.~Mathur, D.~Yarats, and L.~Pinto, ``Watch and match: Supercharging
  imitation with regularized optimal transport,'' \emph{arXiv preprint
  arXiv:2206.15469}, 2022.

\bibitem{Kumar2015}
V.~Kumar and E.~Todorov, ``Mujoco haptix: A virtual reality system for hand
  manipulation,'' in \emph{Humanoids}, 2015.

\bibitem{li2020mobile}
S.~Li, J.~Jiang, P.~Ruppel, H.~Liang, X.~Ma, N.~Hendrich, F.~Sun, and J.~Zhang,
  ``A mobile robot hand-arm teleoperation system by vision and imu,'' in
  \emph{2020 IEEE/RSJ International Conference on Intelligent Robots and
  Systems (IROS)}.\hskip 1em plus 0.5em minus 0.4em\relax IEEE, 2020, pp.
  10\,900--10\,906.

\bibitem{li2019vision}
S.~Li, X.~Ma, H.~Liang, M.~G{\"o}rner, P.~Ruppel, B.~Fang, F.~Sun, and
  J.~Zhang, ``Vision-based teleoperation of shadow dexterous hand using
  end-to-end deep neural network,'' in \emph{2019 International Conference on
  Robotics and Automation (ICRA)}.\hskip 1em plus 0.5em minus 0.4em\relax IEEE,
  2019, pp. 416--422.

\bibitem{megatrack}
S.~Han, B.~Liu, R.~Cabezas, C.~D. Twigg, P.~Zhang, J.~Petkau, T.-H. Yu, C.-J.
  Tai, M.~Akbay, Z.~Wang, A.~Nitzan, G.~Dong, Y.~Ye, L.~Tao, C.~Wan, and
  R.~Wang, ``Megatrack: Monochrome egocentric articulated hand-tracking for
  virtual reality,'' 2020.

\bibitem{hentschel2022steady}
T.~Hentschel and J.~A. Neuh{\"o}fer, ``Steady hands - an evaluation on the use
  of hand tracking in virtual reality training in nursing,'' in
  \emph{Proceedings of the 21st Congress of the International Ergonomics
  Association (IEA 2021)}, N.~L. Black, W.~P. Neumann, and I.~Noy, Eds.\hskip
  1em plus 0.5em minus 0.4em\relax Cham: Springer International Publishing,
  2022, pp. 643--649.

\bibitem{salvato2022predicting}
M.~Salvato, N.~Heravi, A.~M. Okamura, and J.~Bohg, ``Predicting hand-object
  interaction for improved haptic feedback in mixed reality,'' \emph{IEEE
  Robotics and Automation Letters}, vol.~7, no.~2, pp. 3851--3857, 2022.

\bibitem{gharaybeh2019telerobotic}
Z.~Gharaybeh, H.~Chizeck, and A.~Stewart, ``Telerobotic control in virtual
  reality,'' in \emph{OCEANS 2019 MTS/IEEE SEATTLE}, 2019, pp. 1--8.

\bibitem{zhang2018deep}
T.~Zhang, Z.~McCarthy, O.~Jow, D.~Lee, X.~Chen, K.~Goldberg, and P.~Abbeel,
  ``Deep imitation learning for complex manipulation tasks from virtual reality
  teleoperation,'' in \emph{ICRA}, 2018.

\bibitem{MVP}
\BIBentryALTinterwordspacing
I.~Radosavovic, T.~Xiao, S.~James, P.~Abbeel, J.~Malik, and T.~Darrell,
  ``Real-world robot learning with masked visual pre-training,'' 2022.
  [Online]. Available: \url{https://arxiv.org/abs/2210.03109}
\BIBentrySTDinterwordspacing

\bibitem{old_dex_man_1}
V.~Kumar, Y.~Tassa, T.~Erez, and E.~Todorov, ``Real-time behaviour synthesis
  for dynamic hand-manipulation,'' in \emph{2014 IEEE International Conference
  on Robotics and Automation (ICRA)}, 2014, pp. 6808--6815.

\bibitem{old_dex_man_2}
I.~Mordatch, Z.~Popovi\'{c}, and E.~Todorov, ``Contact-invariant optimization
  for hand manipulation,'' in \emph{Proceedings of the ACM
  SIGGRAPH/Eurographics Symposium on Computer Animation}, ser. SCA '12.\hskip
  1em plus 0.5em minus 0.4em\relax Goslar, DEU: Eurographics Association, 2012,
  p. 137–144.

\bibitem{old_dex_man_3}
R.~Deimel and O.~Brock, ``A novel type of compliant and underactuated robotic
  hand for dexterous grasping,'' \emph{The International Journal of Robotics
  Research}, vol.~35, no. 1-3, pp. 161--185, 2016.

\bibitem{old_dex_man_4}
J.~Mahler, M.~Matl, V.~Satish, M.~Danielczuk, B.~DeRose, S.~McKinley, and
  K.~Goldberg, ``Learning ambidextrous robot grasping policies,'' \emph{Science
  Robotics}, vol.~4, no.~26, p. eaau4984, 2019.

\bibitem{Kumar2016}
V.~Kumar, A.~Gupta, E.~Todorov, and S.~Levine, ``Learning dexterous
  manipulation policies from experience and imitation,'' \emph{arXiv}, 2016.

\bibitem{Nagabandi2019}
A.~Nagabandi, K.~Konoglie, S.~Levine, and V.~Kumar, ``Deep dynamics models for
  learning dexterous manipulation,'' \emph{arXiv}, 2019.

\bibitem{Rajeswaran2018}
A.~Rajeswaran, V.~Kumar, A.~Gupta, G.~Vezzani, J.~Schulman, E.~Todorov, and
  S.~Levine, ``Learning complex dexterous manipulation with deep reinforcement
  learning and demonstrations,'' in \emph{RSS}, 2018.

\bibitem{radosavovic2021state}
I.~Radosavovic, X.~Wang, L.~Pinto, and J.~Malik, ``State-only imitation
  learning for dexterous manipulation,'' in \emph{2021 IEEE/RSJ International
  Conference on Intelligent Robots and Systems (IROS)}.\hskip 1em plus 0.5em
  minus 0.4em\relax IEEE, 2021, pp. 7865--7871.

\bibitem{ericsson2022self}
L.~Ericsson, H.~Gouk, C.~C. Loy, and T.~M. Hospedales, ``Self-supervised
  representation learning: Introduction, advances, and challenges,'' \emph{IEEE
  Signal Processing Magazine}, vol.~39, no.~3, pp. 42--62, 2022.

\bibitem{grill2020bootstrap}
J.-B. Grill, F.~Strub, F.~Altch{\'e}, C.~Tallec, P.~Richemond, E.~Buchatskaya,
  C.~Doersch, B.~Avila~Pires, Z.~Guo, M.~Gheshlaghi~Azar, \emph{et~al.},
  ``Bootstrap your own latent-a new approach to self-supervised learning,''
  \emph{NeurIPS}, 2020.

\bibitem{zhang2020mediapipe}
F.~Zhang, V.~Bazarevsky, A.~Vakunov, A.~Tkachenka, G.~Sung, C.-L. Chang, and
  M.~Grundmann, ``Mediapipe hands: On-device real-time hand tracking,'' 2020.

\bibitem{ros}
\BIBentryALTinterwordspacing
{Stanford Artificial Intelligence Laboratory et al.}, ``Robotic operating
  system.'' [Online]. Available: \url{https://www.ros.org}
\BIBentrySTDinterwordspacing

\bibitem{chen2020simple}
T.~Chen, S.~Kornblith, M.~Norouzi, and G.~Hinton, ``A simple framework for
  contrastive learning of visual representations,'' \emph{arXiv preprint},
  2020.

\bibitem{chen2021empirical}
X.~Chen, S.~Xie, and K.~He, ``An empirical study of training self-supervised
  vision transformers,'' in \emph{Proceedings of the IEEE/CVF International
  Conference on Computer Vision}, 2021, pp. 9640--9649.

\bibitem{bardes2021vicreg}
A.~Bardes, J.~Ponce, and Y.~LeCun, ``Vicreg: Variance-invariance-covariance
  regularization for self-supervised learning,'' \emph{arXiv preprint
  arXiv:2105.04906}, 2021.

\bibitem{young2021playful}
S.~Young, J.~Pari, P.~Abbeel, and L.~Pinto, ``Playful interactions for
  representation learning,'' \emph{arXiv preprint arXiv:2107.09046}, 2021.

\end{thebibliography}

\end{document}